\documentclass[conference]{IEEEtran}
\IEEEoverridecommandlockouts
\usepackage{cite}
\usepackage{amsmath,amssymb,amsfonts}
\usepackage{algorithm, algpseudocode, float}
\usepackage{graphicx}
\usepackage{textcomp}
\usepackage{tikz}
\usetikzlibrary{arrows.meta,positioning}
\usepackage{multirow}
\usepackage{xcolor}
\usepackage{comment}
\usepackage{booktabs}

\usepackage{subcaption}
\usepackage{tikz}
\usepackage{multirow}
\usetikzlibrary{arrows.meta}

\def\BibTeX{{\rm B\kern-.05em{\sc i\kern-.025em b}\kern-.08em
    T\kern-.1667em\lower.7ex\hbox{E}\kern-.125emX}}
\begin{document}

\title{Online-Adaptive Fuzzy Cognitive Maps for Neuro-symbolic Financial Forecasting}

\author{
\IEEEauthorblockN{Brandon Rodriguez}
\IEEEauthorblockA{\textit{College of Engineering and Computing} \\
\textit{Florida International University}\\ 
Miami, FL 33199  \\
brodr299@fiu.edu}
\and
\IEEEauthorblockN{Abel Deulofeu}
\IEEEauthorblockA{\textit{College of Engineering and Computing} \\
\textit{Florida International University}\\
Miami, FL 33199 \\
adeul001@fiu.edu}
\and
\IEEEauthorblockN{Olaoluwa Adigun}
\IEEEauthorblockA{\textit{College of Engineering and Computing} \\
\textit{Florida International University}\\
Miami, FL 33199 \\
olaadigu@fiu.edu}
}

\maketitle
\vspace{-0.3in}
\begin{abstract}
We present a neuro-symbolic framework that pairs online-adaptive Fuzzy Cognitive Maps (FCMs) with long short-term memory (LSTM) networks for financial forecasting. 
Financial markets exhibit regime shifts and nonlinear dynamics.
Existing models often trade off one of these key properties: these models capture temporal patterns but hide causal drivers, or they recover causal graphs that cannot adapt post-training. 
We close this gap with two frameworks. 
Framework I constructs an FCM from live dual-source price series and macroeconomic indicators, and adapts its edges online through an error-driven Hebbian rule.
We introduce the Asymmetric Additive Residual Fusion method to seamlessly combine the FCM and LSTM predictions under Framework I.
Framework II constructs a literature-grounded, LLM-augmented FCM where every edge carries a verifiable citation that gives an auditable map of factor interactions. 
On NASDAQ Composite data, Framework I matched the performance of the LSTM baseline on state-directional accuracy and outperformed LSTM by $7.6$ percentage points on the return-sign accuracy. 
Framework I outperformed standalone FCM models across both evaluation metrics. 
The adaptive maps converged to new regimes within $10$--$15$ trading days, and Framework II center-node FCMs scored $68.9\%$ to $85.7\%$ on the state-directional accuracy.
\end{abstract}


\section{Introduction}
\label{sec:introduction}

Various heterogeneous factors and their interactions drive financial markets.
These factors include market volatility, financial crises, firm-specific news, earnings, geopolitical events and so on \cite{maurya2026investor,saravanos2025forecasting, shiller2014speculative,tetlock2007giving,khalil2026ai}.
They operate under incomplete information and behavioral biases.
This contributes to the complexity of modeling such systems.
The time-dependency and change in these factors explain the dynamic nature of financial market modeling \cite{hens2009handbook,chiu2018financial,procacci2023non}.

Traditional econometric frameworks such as autoregressive integrated moving average (ARIMA) \cite{mills2008econometric,idrees2019prediction}, vector autoregression \cite{kilian2017structural}, and linear factor models assume stationarity and linearity \cite{bai2008large}.
But historical financial market data suggests otherwise.
It shows that asset returns exhibit heavy tails, volatility clustering, nonlinear dependencies, and structural breaks that violate the linear and stationary assumptions.
Financial practitioners also need transparent models that are interpretable and reveal causal drivers because of the financial implications of decisions based on such predictions.

Most machine learning models for financial market prediction fail to combine the desirable properties in a single framework: handling heterogeneous data, causal structure, dynamic behavior, and interpretability.
Some combine heterogeneous signals, technical indicators from news and media attention in neural-based models such as LSTMs and transformers to improve prediction performance \cite{htun2023survey,wang2024ensemble,haryono2023transformer}.
Most of them use post-hoc tools like SHAP and LIME to achieve explainability \cite{muhammad2024explainable}.
Neural-based methods excel at capturing long-range dependencies but they lack explicit causal capability and interpretability.
Other prior works extract causal graphs from financial news \cite{DBLP:journals/eswa/DengLY24} and learn causal representations for robustness \cite{DBLP:journals/tkde/XuCL26}.
However, their causality is acyclic and static, interpretability is external to the predictor, and inputs or functional form are restricted.

\begin{figure*}[!t]
\centering
\includegraphics[width=0.9\linewidth]{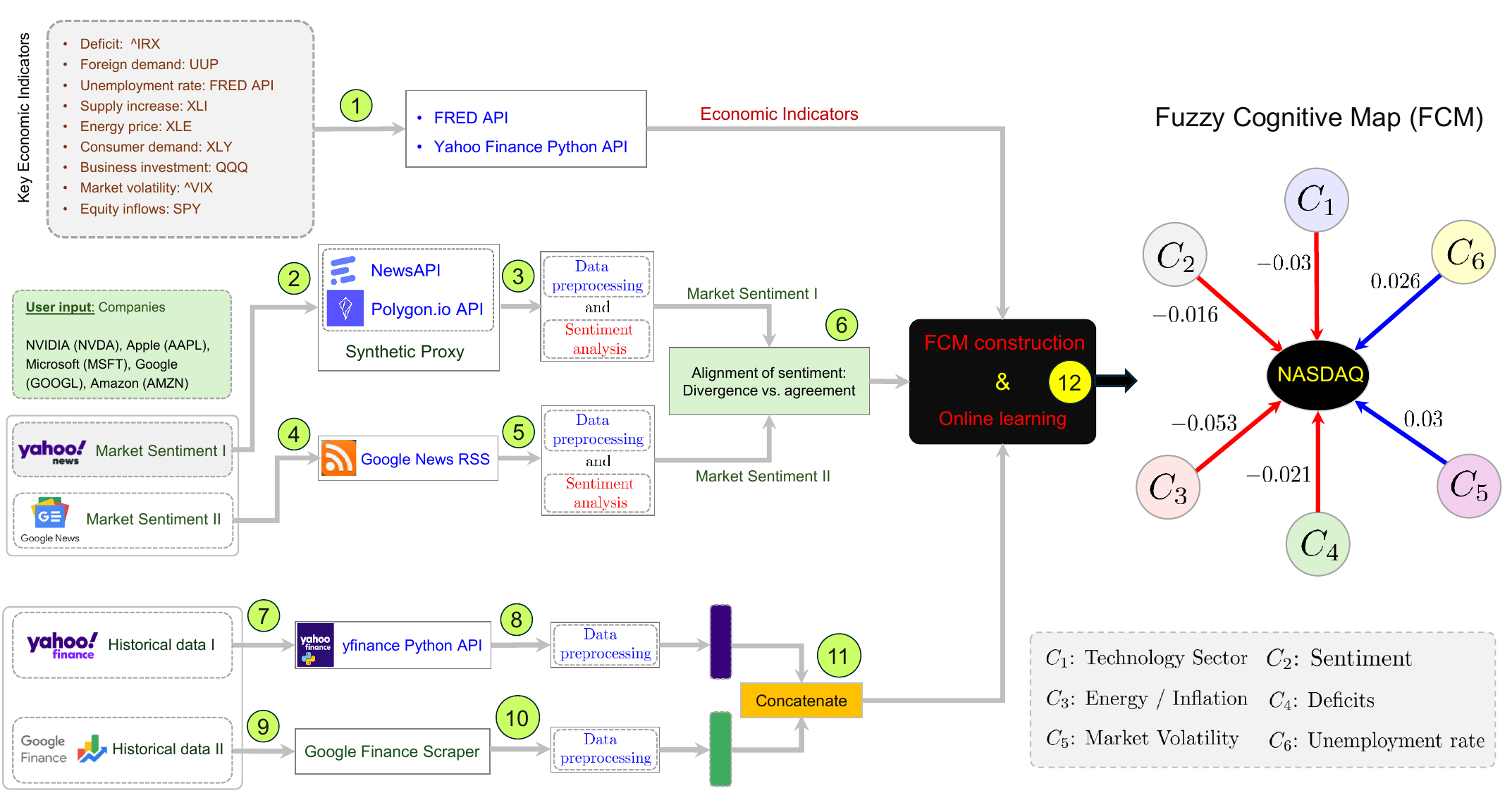}
\caption{\footnotesize Data pipeline and online FCM construction for Framework I:
The system ingests three input streams and fuses them into an online-adapting fuzzy cognitive map.
Step (1) pulls key economic indicators through the FRED API and the \texttt{yFinance} Python API.
For news sentiment, the user specifies target companies (NVDA, AAPL, MSFT, GOOGL, AMZN); steps (2--3) query NewsAPI and the Polygon.io API as a synthetic proxy for Yahoo News, then preprocess the text and run sentiment analysis to produce Market Sentiment I.
Steps (4--5) scrape Google News RSS and apply the same preprocessing and sentiment analysis to produce Market Sentiment II.
Step (6) aligns the two sentiment streams and measures their divergence versus agreement. For price history, steps (7--8) retrieve Yahoo Finance data through the Yahoo Finance API and preprocess it.
Steps (9--10) retrieve Google Finance data through a scraper and preprocess it, and step (11) concatenates the two price streams.
Step (12) constructs the FCM and performs online learning, combining the economic indicators, aligned sentiment, and concatenated prices while updating edge weights as new data arrives.
The right panel shows the resulting fuzzy cognitive map after online adaptation on NASDAQ Composite data (2020--2025). Blue arrows denote positive causal influence; red arrows denote negative influence. Edge labels show the final adapted weights $w_{i,\texttt{NASDAQ}}$. Energy price increase carries the strongest suppressive effect ($-0.053$), while low unemployment and elevated market volatility carry positive influence on NASDAQ.}
\vspace{-0.2in}
\label{fig:fcm_framework_1}
\end{figure*}

\begin{figure*}[t]
\centering
\includegraphics[width=0.9\linewidth]{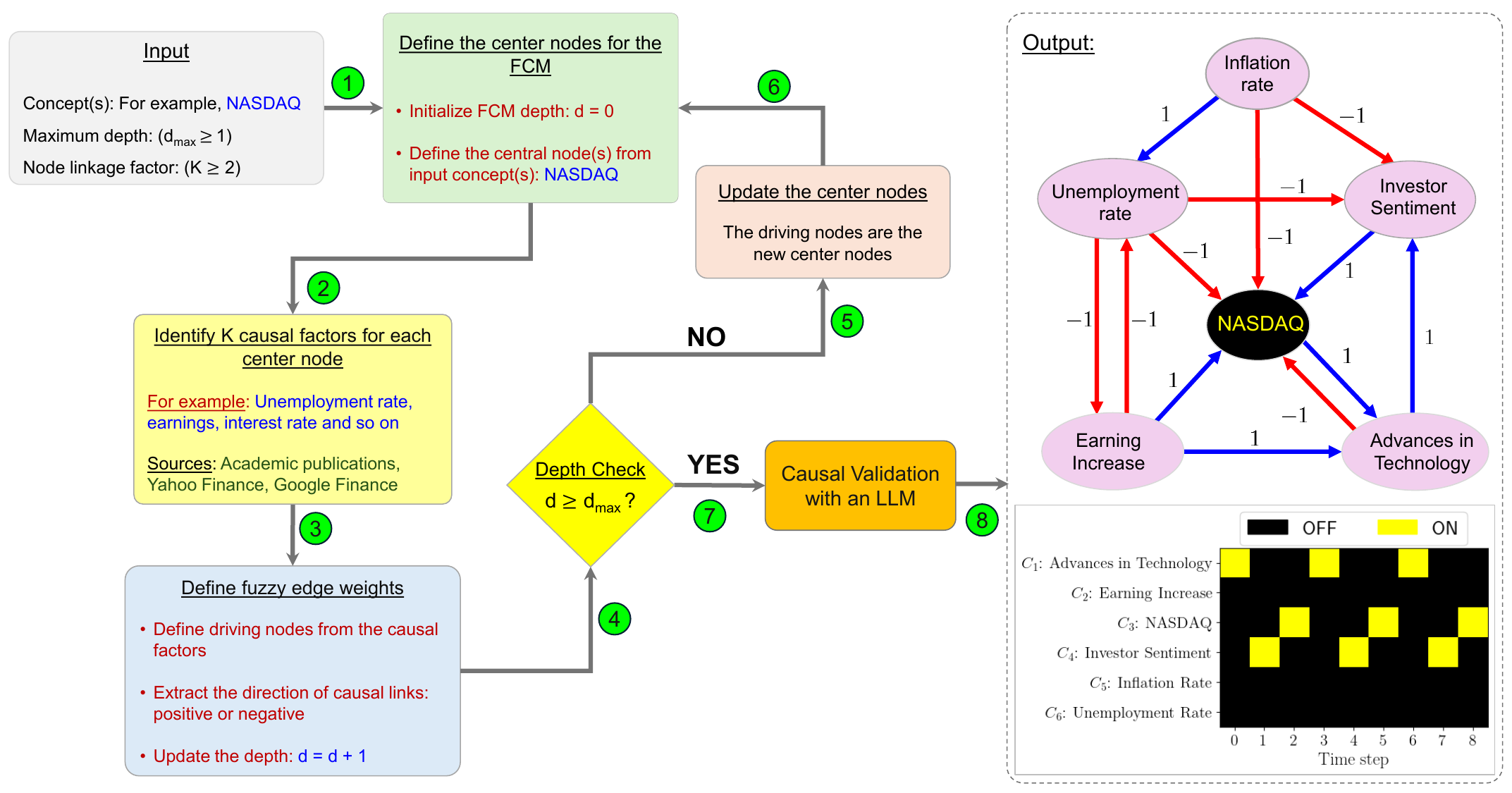}
\caption{\footnotesize Iterative, literature-grounded FCM construction for Framework II:
This expands a fuzzy cognitive map outward from one or more seed concepts until it reaches a maximum depth, then validates the result with an LLM.
Given an input concept (for example, NASDAQ), a maximum depth $d_\texttt{max}$, and a node linkage factor $K \geq 2$, step (1) sets the seed as the central node and initializes the depth to $d = 0$.
Step (2) identifies $K$ causal factors for each center node, such as unemployment rate, earnings, and interest rate, and draws them from academic publications and financial data sources (Yahoo Finance and Google Finance).
Step (3) defines the fuzzy edge weights: it designates the causal factors as driving nodes, extracts the sign of each causal link as positive or negative, and increments the depth to $d = d + 1$.
Step (4) checks whether $d \geq d_\texttt{max}$.
If not, step (5) promotes the current driving nodes to center nodes and step (6) returns to step (1) to expand the next layer.
Once $d \geq d_\texttt{max}$, step (7) validates the causal links with an LLM and step (8) produces the final map.
The output panel shows the constructed FCM centered on NASDAQ, where blue edges denote positive influence, red edges denote negative influence, and the labels give the signed edge weights.
The raster below shows the map's dynamics across time steps 0 to 8.}
\vspace{-0.1in}
\label{fig:fcm_framework_2}
\end{figure*}

Fuzzy Cognitive Maps (FCMs) provide an interpretable graph-based framework for modeling complex systems such as financial markets.
FCMs are directed graphs that model the causal relationships in a system \cite{Kosko86,panda2025causal}.
Their nodes denote concepts, and their signed, weighted edges quantify the direction and strength of each causal link.
FCMs admit feedback loops and iterate to produce dynamic trajectories rather than static outputs for acyclic causal graphs \cite{DBLP:journals/tsmc/StyliosG04}.
FCM predictions trace directly to the signed edges that drove them and build explainability into the mechanism rather than post-hoc recovery.
This pairing of causal structure, feedback dynamics, and intrinsic transparency in one model has driven FCM adoption across complex decision domains \cite{DBLP:journals/tfs/PapageorgiouS13,DBLP:journals/air/FelixNFFVB19}.

Some prior works have applied FCMs for financial analysis ranging from expert-built maps of stock-market dynamics to interval-valued maps \cite{DBLP:conf/fuzzIEEE/KoulouriotisDE01,hajek2017learning}.
The recent work CausalSeqGNN is a hybrid model that gives FCM nodes Fama-French-inspired fuzzy state vectors, learns the map with a genetic algorithm, and couples it with a graph neural network for cross-sectional forecasting \cite{DBLP:journals/tfs/YuL25}.
These methods either extract these maps from experts or learn them offline.
After training, they hold the map fixed.
This leaves two questions open. How can we build FCMs from heterogeneous multi-source feeds and literature-grounded LLM knowledge? And how can we adapt them online as the market moves?

We present two frameworks for neuro-symbolic modeling with FCMs to address these questions.
Framework I is a hybrid model that uses an FCM (or multiple FCMs) to extract and merge explainable features with LSTM neural processing for prediction.
This framework uses live market feeds to extract useful knowledge and features.
This framework extracts information from financial news sentiment and key macroeconomic indicators to build a combined FCM or two FCMs (one for market sentiment and the other for sector-specific information).
We introduce the Asymmetric Additive Residual Fusion method to combine dual FCMs with deep LSTM neural networks for prediction.
Figure~\ref{fig:fcm_framework_1} shows the FCM pipeline for Framework I and the resulting causal map.

Framework II  fuses two knowledge sources, peer-reviewed literature and LLM knowledge, under human oversight, and produces a transparent, verifiable map of how financial factors interact rather than a forecast.
Figure~\ref{fig:fcm_framework_2} shows the FCM pipeline for Framework II. 
Our work makes three main contributions: (1) We introduce an online FCM adaptation method. (2) We build a neuro-symbolic model combining FCMs and deep LSTM networks using the asymmetric additive residual fusion method. (3) We construct a literature-grounded, LLM-augmented FCM with per-edge citations. %

On NASDAQ data, Framework I matches LSTM state-directional  accuracy ($78.8\%$) while outperforming standalone LSTM by $7.6$ percentage points in return-sign accuracy $(52.0\%)$ and pure FCMs by $15.6$ percentage points.
The adaptive FCMs converge to new market regimes within $10$--$15$ trading days. 
The literature-grounded Framework~II maps reach $68\%$--$85\%$ state-directional accuracy. 

\section{Background and Notation}
\subsection{Fuzzy Cognitive Map}
An FCM is a directed graph $\mathcal{G}=(\mathcal{N},\mathcal{E})$ with node set $\mathcal{N}$ and edge set $\mathcal{E}$ with each edge weight in $[-1,1]$.
The value of $w_{ij}$ indicates the strength  and direction of the causal link from the $i^{th}$ node $C_i$ to the $j^{th}$ node $C_j$.
Positive weights indicate excitatory influence, negative weights mean inhibitory influence, and zero weights mean no influence\cite{panda2025agentic}.

The edge weight of an adaptive FCM depends on time step $t$.
We define $w_{ij}^{(t)}$ as the edge weight at time $t$.
This generalizes to a static FCM, where $w_{ij}^{(t)}$ is the same for all $t$.
The state vector ${\bf C}^{(t)} = \big{[}C_1^{(t)},\dots,C_N^{(t)}\big{]}^\textsf{T}$ represents the state of FCM nodes at time $t$, where $C_i^{(t)}$ is the value of node $C_i$ at time $t$.
The state evolution is
\begin{equation}\label{eq:state_evolution}
C_j^{(t+1)} = f\Big{(} \sum_{i=1}^N w_{ij}^{(t)} C_i^{(t)} + b_j^{(t)}\Big{)}
\end{equation}
where $f(\cdot)$ is the logistic sigmoid $\sigma(x)=1/(1+e^{-cx})$, $c \geq 0$, and $b_j^{(t)}$ represents the exogenous bias (e.g., market sentiment) at time $t$.

\subsection{Long Short-Term Memory (LSTM) Network}
A long short-term memory (LSTM) network is a recurrent neural network that processes a sequence one step at a time while maintaining a cell state.
It uses input, forget, and output gates to control what information it stores, discards, and exposes at each step \cite{schmidhuber1997long}.
These gates let the gradient flow across many steps without vanishing, so the LSTM learns long-range temporal dependencies that a plain recurrent neural network cannot.
The time window parameter $T$ of an LSTM sets the look-back length. This means that the network reads the most recent $T$ consecutive time steps as one input sequence and produces its prediction from that window.
Each window holds $T$ successive 16-node FCM concept state vectors for the LSTM input in Framework I.

\subsection{Large Language Model (LLM) Domain Knowledge}
A Large Language Model (LLM) is a deep neural network trained on extensive text corpora. 
These corpora encode broad domain knowledge, including financial concepts, market mechanisms, and economic relationships.
We draw on this knowledge to propose and verify causal links between concept nodes in financial domain FCMs. 
The LLM cites supporting source(s) for each link because LLM outputs are not always reliable.
The LLM provides verifiable literature citations for every causal edge in the graph to mitigate potential model hallucinations and ensure full auditability.

\section{Data Pipeline}

We combine multi-source financial data and domain knowledge into computational neuro-symbolic models. 
In this section, we detail the data sources and preprocessing methods for both frameworks.

\subsection{Framework I}
We combine structured data including historical price and macroeconomic data, and unstructured market sentiment data.

\subsubsection{Historical Price Data}
This captures the historical price of the target company or index fund over the $T$ steps before time $t$.
It also captures the prices of other companies or funds or values of indicators that drive the target.
We collect historical market data using the \texttt{yfinance} Python library for Yahoo Finance, a custom Python scraper for Google Finance, or a combination of both (Figure~\ref{fig:fcm_framework_1}).
We combine both sources when individual providers lack complete coverage across tickers, timeframes, or data fields. 
This dual-sourcing strategy eliminates data gaps and maintains continuous pipeline execution during API rate limits or downtime.

We scaled all price data values to log-returns $r_t$:
\begin{align}
r_t = \ln\frac{P_t}{P_{t-1}}.
\end{align}
The pipeline uses rolling min-max normalization over a 252-day window to scale all features to $[0,1]$:
\begin{align}\label{eqn:preprocess_price_features}
\tilde{x}_t = \frac{x_t - \min(x_{t-252:t})}{\max(x_{t-252:t}) - \min(x_{t-252:t}) + 10^{-12}}.
\end{align}
{\color{black}{Here $x_t \in \mathbb{R}$ is the raw feature value at time $t$ (including log-returns, macroeconomic indicator values, or indicator signals) and the 252-day window represents one trading year of daily observations.}}

\subsubsection{Key Economic Indicators}

We incorporate market and macroeconomic proxies that capture the multi-horizon conditions driving financial markets. Market volatility (\texttt{vix}, \texttt{vol\_surge}) reflects near-term risk and stress, while short-term interest rates (\texttt{rate3m}), real yield momentum (\texttt{real\_yield\_diff}), inflation expectations (\texttt{breakeven5y}), and unemployment rates (\texttt{unemp}) set the medium-to-long-term regime in which asset prices evolve. We include these proxies because a price-only model sees the symptom but not the underlying economic cause; incorporating them enables the FCM to explain why the market reacts rather than merely predict that it does.

We track these 16 concept proxies across five categories: FRED \texttt{UNRATE} for unemployment (\texttt{unemp}), FRED \texttt{VIXCLS} for volatility (\texttt{vix}, \texttt{vol\_surge}), FRED \texttt{DGS3MO} for 3-month Treasury yields (\texttt{rate3m}), FRED \texttt{T5YIE} for 5-year breakeven inflation (\texttt{breakeven5y}), sector ETFs such as \texttt{XLE} for energy (\texttt{energy}) and \texttt{QQQ}/\texttt{ONEQ} for market concentration (\texttt{breadth}, \texttt{tech\_rotation}), alongside tech leadership vectors (\texttt{tech}, \texttt{ai\_leadership}) and technical momentum (\texttt{rsi}, \texttt{m5}, \texttt{m20}).
Step (1) in Figure~\ref{fig:fcm_framework_1} shows the key economic indicator extraction pipeline.

\subsubsection{Market Sentiment}
Framework I fetches news headlines daily for the target companies from Yahoo News, Google News, or the combination of the two.
It uses the NewsAPI to extract from Yahoo News and falls back to the Polygon.io API if that fails.
For Google News, we use the Google RSS feed.
The pipeline preprocesses each stream and runs sentiment analysis on the headlines to produce two independent sentiment signals.
The pipeline uses the Loughran-McDonald lexicon \cite{loughran2011liability} to compute a raw polarity score $S_t$:
\begin{align}
S_t &= \frac{N_{\text{pos}} - N_{\text{neg}}}{N_{\text{total}}}
\end{align}
{\color{black}{where $N_{\text{pos}}$ is the count of Loughran-McDonald positive-sentiment words, $N_{\text{neg}}$ is the count of negative-sentiment words, and $N_{\text{total}}$ is the total word count across all headlines on day $t$.
Therefore we have $S_t \in [-1, 1]$.
We smooth the raw score using an exponential moving average with a decay coefficient $\delta = 0.8$
\begin{align}\label{eqn:running_score}
R_t = \delta\cdot R_{t-1} + (1-\delta)\cdot S_t,
\end{align}
delay it by one trading day to avoid look-ahead bias, and bound it with
$\tilde{S}_t = \tanh(5\cdot R_{t-1})$.}}
In the dual-sentiment configuration, the pipeline processes Yahoo News and Google News independently, and aligns both signals to measure their agreement or divergence before feeding them to the FCM (Figure~\ref{fig:fcm_framework_1}).

\subsection{Framework~II: Data Pipeline}
Framework II constructs its FCM from two static knowledge sources rather than live market feeds.
Causal nodes and edge signs are drawn from peer-reviewed literature, and an LLM proposes additional links.
A cited justification accompanies each link and a human reviewer can verify it.
No price or sentiment data is ingested at runtime.
The FCM is fixed after the literature-grounding and LLM validation steps.
Figure~\ref{fig:fcm_framework_2} shows the pipeline for constructing the FCM.
We used the GPT-4 LLM for this work. 

\section{Methodology}

\subsection{Framework I}
This subsection explains the design of Framework I.
This includes the FCM construction, its online adaptation, and the asymmetric additive residual hybrid fusion for the FCM-LSTM ensemble.\\

\subsubsection{Fuzzy Cognitive Map Construction}
The state vector evolution follows from Equation (\ref{eq:state_evolution}),
where $b_j^{(t)}$ is the exogenous bias (e.g., the smooth sentiment signal $\tilde{S}_t$ injected into the NASDAQ node).
For the baseline, initial weights are derived from a $\tau$-day sign-correlation:
\begin{align}\label{eqn:weight_init}
w_{ij} = \tanh\!\left(\frac{1}{\tau}\sum_{k=1}^{\tau}
    \operatorname{sign}(\Delta C_{i,k})\cdot \operatorname{sign}(\Delta C_{j,k})\right),
\end{align}
where $\Delta C_{i,k} = C_i^{(k)} - C_i^{(k-1)}$ is the one-day change in the normalized state of concept node $i$.
The product $\operatorname{sign}(\Delta C_{i,k})\cdot\operatorname{sign}(\Delta C_{j,k}) \in \{-1,+1\}$ counts co-directional moves versus counter-directional moves.
The model averages over $\tau$  trading days and applies the hyperbolic tangent $(\tanh)$ to map the real line to $[-1,1]$ and give the initial signed causal weight.
Node activation $A_i(t)$ at time $t$ for the $i^{th}$ stock price uses the sigmoid of normalized returns
\begin{equation}
A_i(t) = \frac{1}{1+e^{-\alpha \cdot r_{i,t}}}
\end{equation}
where $\alpha > 0$ and $r_{i,t}$ is the log-return for node $i$ at time $t$.

{Technical momentum and market microstructure nodes complement the economic  indicators.
The 14-day Relative Strength Index
\begin{align}\label{eqn:rsi}
C_{\texttt{RSI}}(t) = 1 - \frac{1}{1 + \bar{g}_{14}/\bar{l}_{14}} \;\in [0,1]
\end{align}
where $\bar{g}_{14}$ and $\bar{l}_{14}$ are the respective 14-day average gain and loss.
$C_{\texttt{RSI}}>0.7$ indicates overbought regime and $C_{\texttt{RSI}}<0.3$ signals oversold regime.
We define the short-price momentum $C_{\texttt{M5}}(t)$ and the medium-term price momentum $C_{\texttt{M20}}(t)$ as follows:
\begin{align}\label{eqn:momentum}
C_{\texttt{M5}}(t) &= \tanh\!\left(20\cdot r_{5d,t}\right) \\
C_{\texttt{M20}}(t) &= \tanh\!\left(10\cdot r_{20d,t}\right)
\end{align}
where $r_{5d,t}$ and $r_{20d,t}$ are the 5-day and 20-day cumulative log-returns.
We define four market microstructure concepts to capture market concentration, sector rotation, and volatility acceleration.
They are:
\begin{align}
C_{\texttt{BREADTH}}(t) &= \text{norm}\left( \frac{\text{Price}_{\texttt{QQQ},t}}{\text{Price}_{\texttt{ONEQ},t}} \right) \\
C_{\texttt{ROTATION}}(t) &= \text{norm}\left( r_{\texttt{QQQ},t} - r_{\texttt{XLE},t} \right) \\
C_{\texttt{VSURGE}}(t) &= \text{norm}\left( \frac{\texttt{VIX}_t}{\text{SMA}_{20}(\texttt{VIX})_t} \right) \\
C_{\texttt{REAL\_YIELD}}(t) &= \text{norm}\left( \Delta \texttt{DGS3MO}_t - \Delta \texttt{T5YIE}_t \right)
\end{align}
where $\text{norm}(\cdot)$ denotes the rolling $252$-day min-max normalization into $[0,1]$ defined in Equation~(\ref{eqn:preprocess_price_features}).}

\subsubsection{Online Adaptation}
Unlike static FCMs, we update weights at every time step using prediction error on the target node (NASDAQ). Algorithm~\ref{alg:fcm_adapt} details the online adaptation rule.
\begin{algorithm}[!t]
\caption{: FCM Online Edge Weight Adaptation}
\label{alg:fcm_adapt}
\begin{algorithmic}[1]
\Require Realized return $r_t$; learning rate $\eta${\color{black}{$=0.05$}}; {\color{black}{momentum $\beta=0.9$}}; predicted state $\hat{C}_{\texttt{NASDAQ}}^{(t)}$; current node states $C_j^{(t)}$
\State Compute prediction error: $\epsilon_t = r_t - \hat{C}_{\texttt{NASDAQ}}^{(t)}$
\State Calculate derivative: $\delta = \epsilon_t \cdot \hat{C}_{\texttt{NASDAQ}}^{(t)} \cdot (1 - \hat{C}_{\texttt{NASDAQ}}^{(t)})$ {\color{blue}{\Comment{\texttt{Logistic derivative}}}}
\For{each incoming edge $w_{i,\texttt{NASDAQ}}$}
    \State {\color{black}{$v_i \leftarrow \beta\cdot v_i + \eta\cdot\delta\cdot C_i^{(t)}$}} \hfill {\color{blue}{\Comment{\texttt{Momentum accumulation}}}}
    \State {\color{black}{$w_{i,\texttt{NASDAQ}} \leftarrow \operatorname{clip}\!\left(w_{i,\texttt{NASDAQ}} + v_i,\;-1,\;1\right)$}}
\EndFor
\State {\color{black}{Verify spectral radius $\rho(\mathbf{W}) \!=\! \max_k|\lambda_k(\mathbf{W})| < 1$ for convergence. %
}}
\end{algorithmic}
\end{algorithm}
This Hebbian-style update strengthens edges that reduce prediction error. The dual pipelines (Yahoo and Google) adapt independently.
The sentiment nodes are exogenous inputs that do not receive incoming updates but influence other nodes.
Figure~\ref{fig:fcm_adapt} compares the FCMs for the two pipelines over the same period.

\subsubsection{FCM-LSTM Ensemble}
While FCMs capture concept interactions, deep recurrent networks excel at learning long-range temporal dependencies.
We used an LSTM recurrent neural network with $50$ units and dropout value $0.2$ and combined it with the adaptive FCM.
The network processes input sequences of length $T = 30$ trading days, where each time step collects the 16-node concept state vector.
We use the asymmetric additive residual fusion method to combine the FCMs with the LSTM:
\begin{align}
\hat{y}_t^{\texttt{ensemble}} = \texttt{clip}\!\left( \hat{y}_t^{\text{LSTM}} + \alpha(\Delta_t^{\text{FCM}}) \cdot \Delta_t^{\text{FCM}}, \, 0, \, 1 \right)
\end{align}
where $\Delta_t^{\text{FCM}} = \frac{1}{2}\left(\hat{y}_t^{\text{FCM-A}} + \hat{y}_t^{\text{FCM-B}}\right) - 0.5$ represents the zero-centered FCM directional steering vector, and the asymmetric multiplier $\alpha(\cdot)$ accounts for market skewness:
\begin{align}
\alpha(\Delta_t^{\text{FCM}}) = \begin{cases}
\alpha_{\text{up}} & \text{if } \Delta_t^{\text{FCM}} > 0 \quad \footnotesize{(\texttt{Bullish signal})} \\
\alpha_{\text{down}} & \text{if } \Delta_t^{\text{FCM}} \le 0 \quad \footnotesize{(\texttt{Bearish signal})}
\end{cases} 
\end{align}
where $\alpha_{\text{up}}, \alpha_{\text{down}} \in [0, 0.5]$ to preserve neural calibration.

\subsection{Framework II}

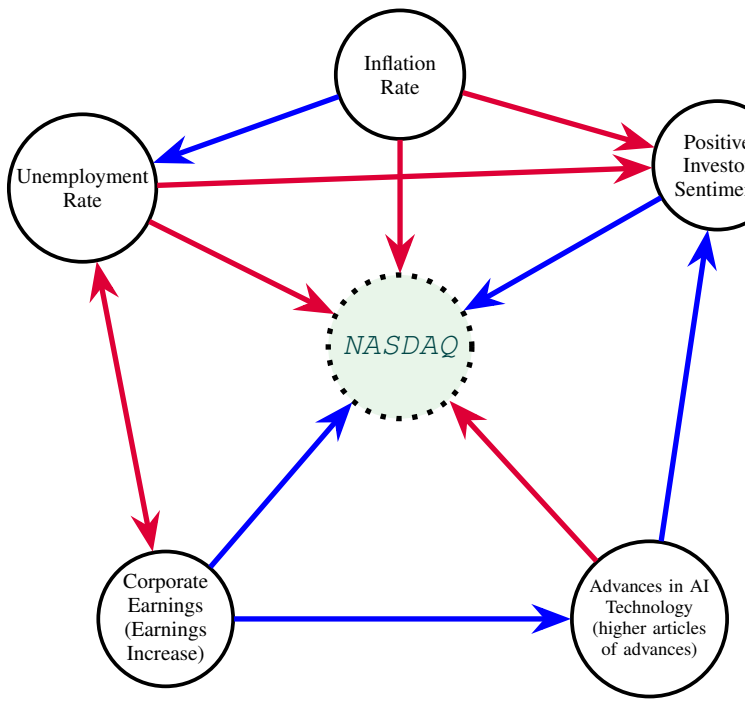
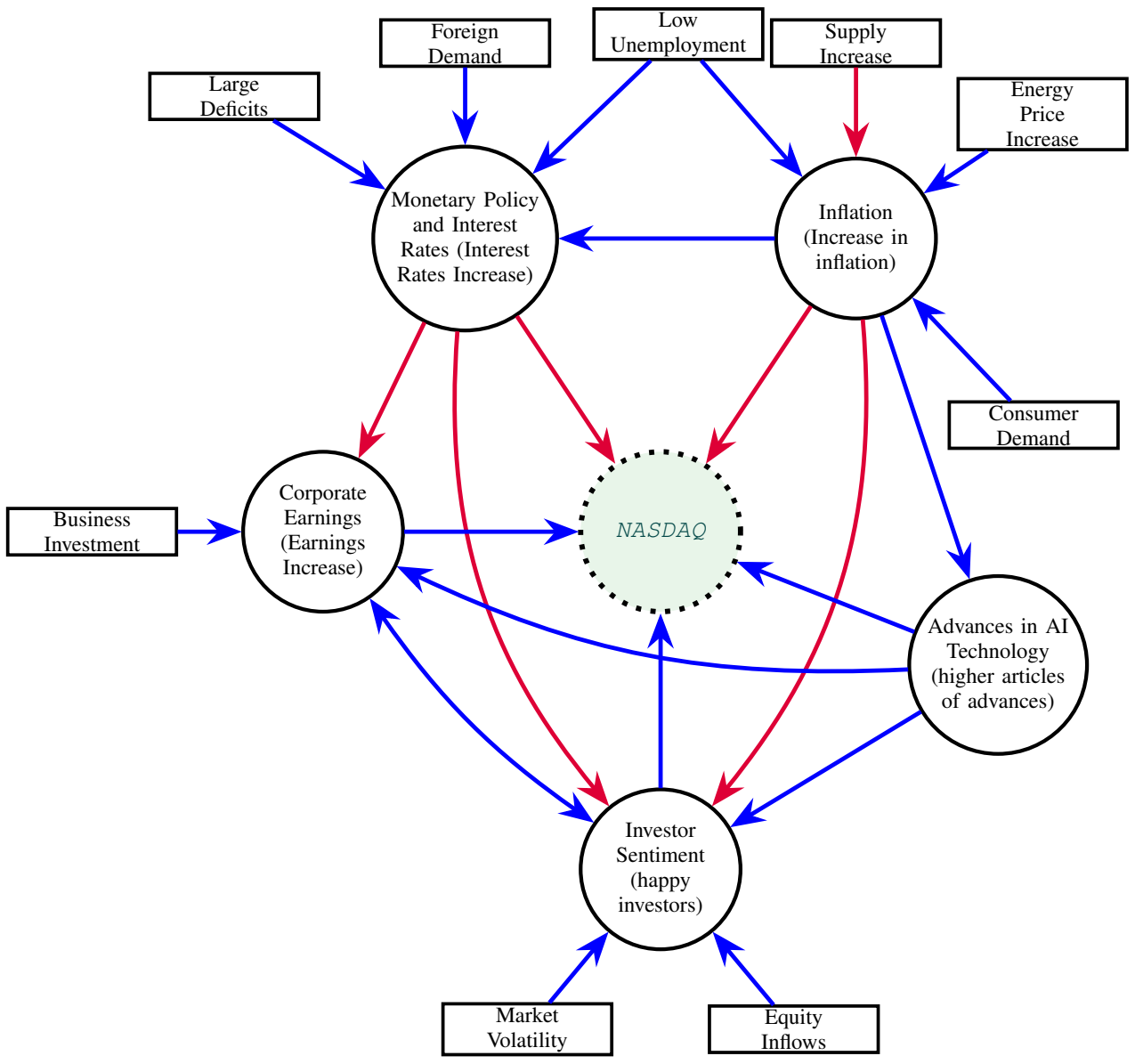
\begin{figure}[t]
\centering
\definecolor{posteal}{RGB}{0,0,255}
\definecolor{negred}{RGB}{222,0,54}
\definecolor{centerfill}{RGB}{232,245,233}

\begin{subfigure}[t]{0.5\textwidth}
\centering
\begin{tikzpicture}[
    scale=0.5, transform shape,
    >={Stealth[length=2.6mm]},
    outer/.style={circle, draw=black, line width=0.7pt, fill=white,
                  minimum size=1.7cm, align=center, font=\footnotesize, inner sep=1.5pt},
    centernode/.style={circle, draw=black, dotted, line width=1pt,
                  fill=centerfill, minimum size=1.9cm, align=center,
                  font=\footnotesize\itshape, text=teal!55!black},
    poslink/.style={->, posteal, line width=1pt},
    neg/.style={->, negred, line width=1pt},
    negbi/.style={<->, negred, line width=1pt},
]
\node[centernode] (nasdaq) at (0,0) {\large \texttt{NASDAQ}};
\node[outer] (infl)  at (-0.0, 3.6)  {Inflation\\ Rate};
\node[outer] (inv)   at ( 4.2, 2.4)  {Positive \\ Investor\\ Sentiment};
\node[outer] (unemp) at (-4.2, 2.1)  {Unemployment\\ Rate};
\node[outer] (corp)  at (-3.1,-3.6)  {Corporate\\Earnings\\(Earnings\\Increase)};
\node[outer, font=\scriptsize] (ai) at (3.3,-3.6)
      {Advances in AI\\Technology\\(higher articles\\of advances)};
\draw[neg]     (infl)  -- (nasdaq);
\draw[neg]     (unemp) -- (nasdaq);
\draw[poslink] (inv)   -- (nasdaq);
\draw[poslink] (corp)  -- (nasdaq);
\draw[neg]     (ai)    -- (nasdaq);
\draw[neg]     (infl)  -- (inv);
\draw[poslink] (infl)  -- (unemp);
\draw[neg]     (unemp) -- (inv);
\draw[negbi]   (unemp) -- (corp);
\draw[poslink] (corp)  -- (ai);
\draw[poslink] (ai)    -- (inv);
\end{tikzpicture}
\caption{\small ${d_{\texttt{max}}} = 1$ }
\label{fig:fcm-nasdaq}
\end{subfigure}

\vspace{1.0\baselineskip}

\begin{subfigure}[t]{0.5\textwidth}
\centering
\begin{tikzpicture}[
    scale=0.66, transform shape,
    >={Stealth[length=2.6mm]},
    main/.style={circle, draw=black, line width=0.8pt, fill=white,
                 minimum size=1.8cm, align=center, font=\scriptsize, inner sep=1pt},
    centernode/.style={circle, draw=black, dotted, line width=1.1pt,
                 fill=centerfill, minimum size=1.8cm, align=center,
                 font=\footnotesize\itshape, text=teal!55!black},
    leaf/.style={rectangle, draw=black, line width=0.7pt, fill=white,
                 align=center, font=\scriptsize, inner sep=1pt,
                 minimum width=1.9cm},
    poslink/.style={->, posteal, line width=0.9pt},
    neg/.style={->, negred, line width=0.9pt},
    posbi/.style={<->, posteal, line width=0.9pt},
]
\node[centernode] (nas) at (-0.8,0)      {\texttt{NASDAQ}};
\node[main] (mp)  at (-3.0, 3.3) {Monetary Policy\\and Interest\\Rates (Interest\\Rates Increase)};
\node[main] (inf) at (1.4, 3.3) {Inflation\\(Increase in\\inflation)};
\node[main] (ce)  at (-4.6, 0.0) {Corporate\\Earnings\\(Earnings\\Increase)};
\node[main] (ai)  at (3.0,-1.5) {Advances in AI\\Technology\\(higher articles\\of advances)};
\node[main] (is)  at (-0.8,-3.8) {Investor\\Sentiment\\(happy\\investors)};
\node[leaf] (ld) at (-5.6, 4.9) {Large\\Deficits};
\node[leaf] (fd) at (-3.0, 5.5) {Foreign\\Demand};
\node[leaf] (lu) at (-0.6, 5.6) {Low\\Unemployment};
\node[leaf] (si) at (1.4, 5.5) {Supply\\Increase};
\node[leaf] (ep) at (3.5, 4.7) {Energy\\Price\\Increase};
\node[leaf] (cd) at (3.4, 1.2) {Consumer\\Demand};
\node[leaf] (bi) at (-7.2, 0.0) {Business\\Investment};
\node[leaf] (mv) at (-2.3,-5.6) {Market\\Volatility};
\node[leaf] (ei) at ( 0.7,-5.6) {Equity\\Inflows};
\draw[poslink] (ld) -- (mp);
\draw[poslink] (fd) -- (mp);
\draw[poslink] (lu) -- (mp);
\draw[poslink] (lu) -- (inf);
\draw[neg]     (si) -- (inf);
\draw[poslink] (ep) -- (inf);
\draw[poslink] (cd) -- (inf);
\draw[poslink] (bi) -- (ce);
\draw[poslink] (mv) -- (is);
\draw[poslink] (ei) -- (is);
\draw[poslink] (inf) -- (mp);
\draw[neg]     (mp)  -- (nas);
\draw[neg]     (mp)  -- (ce);
\draw[neg]     (mp)  to[bend right=22] (is);
\draw[neg]     (inf) -- (nas);
\draw[neg]     (inf) to[bend left=22]  (is);
\draw[poslink] (inf) -- (ai);
\draw[poslink] (ce)  -- (nas);
\draw[posbi]   (ce)  to[bend right=10] (is);
\draw[poslink] (ai)  -- (nas);
\draw[poslink] (ai)  -- (is);
\draw[poslink] (ai)  to[bend left=14]  (ce);
\draw[poslink] (is)  -- (nas);
\end{tikzpicture}
\caption{${d_{\texttt{max}}} = 2$}
\label{fig:fcm-nasdaq-full}
\end{subfigure}
\caption{\footnotesize  Framework II (FCM Extraction): 
These are center-node FCMs for the NASDAQ index using Algorithm~\ref{alg:fcm-framework-2} with secondary factor nodes (circles) feeding the primary concepts (rectangle).
(a) The maximum depth is $d_{\texttt{max}} = 1$ so the FCM added a layer of factors to the center-node.
(b) The maximum depth is $d_{\texttt{max}} = 2$ so the FCM added two layers of factors to the center-node.
}
\vspace{-0.1in}
\label{fig:fcm-nasdaq-merged}
\end{figure}

{
We present the FCM extraction and construction methodology for Framework~II. 
This framework initializes the concept map from a target central node up to a specified graph depth, $d_{\max} \ge 1$. 
Every node must link to at least one neighboring concept to prevent isolated nodes and preserve connectivity.
We constrain all node activation states to $[0, 1]$ with the rolling min-max normalization. 
Similarly, we constrain causal edge weights to $[-1, 1]$ with a hyperbolic tangent activation function to capture both positive and negative relationships. Algorithm~\ref{alg:fcm-framework-2} shows the pseudocode for this framework.

To grow outer concept nodes beyond the central target, we identify key market drivers through literature research. 
We query academic databases (e.g., AI-powered Google Scholar Lab) using standardized prompts such as: \textit{``What are the $X$ most critical factors affecting [node]?''} Following node identification, we establish directional causal links based on documented economic relationships. 
A key design challenge involves quantifying edge magnitudes.
While empirical literature confirms directional polarities (e.g., the inverse relationship between unemployment and inflation), exact numerical weights are rarely stated explicitly. 
We resolve this by assigning polarity-aligned initial weights bounded in $[-1, 1]$.

Once we construct the initial graph topology, we pass the network to a specialized Large Language Model (LLM) tasked with evaluating node validity and edge completeness. The LLM validates every node-edge attribution against published academic literature and provides human-verifiable citations for each link. In practice, this LLM verification step refined graph topology by identifying previously overlooked cross-sector edges and reinforcing existing causal pathways.
Figure~\ref{fig:fcm-nasdaq-merged} shows how the FCM extraction for Framework II works.
}

\begin{algorithm}[!t]
\caption{: Iterative FCM extraction (Framework II)}
\label{alg:fcm-framework-2}
\begin{algorithmic}[1]
\Require Target depth $d_{\texttt{max}}$, node linkage factor $K$
\Ensure Connected FCM with $a_i \in [0,1]$ and $w_{ij} \in [-1,1]$
\State Initialize center node $\mathcal{C}$ {\color{blue}{\Comment{\texttt{Broad market index, e.g.\ NASDAQ}}}}
\State $\mathcal{F} \gets \mathcal{C}$ {\color{blue}{\Comment{\texttt{Frontier: nodes awaiting expansion}}}}
\State $d \gets 0$ {\color{blue}{\Comment{\texttt{Current depth}}}}
\Repeat
    \State $\mathcal{F}' \gets \varnothing$
    \ForAll{$n \in \mathcal{F}$}
        \State Identify $K$ direct causal factors $\{f_1, \dots, f_K\}$ of $n$
        \ForAll{$f_k \in \{f_1, \dots, f_K\}$}
            \State $s(n,fk) \gets$ Polarity score from literature
            \State $w_{n,f_k} \gets \tanh(s(n,fk)) \in [-1,1]$ {\color{blue}{\Comment{\texttt{Fuzzy edge}}}}
            \State normalize $a_{f_k} \in [0,1]$
            \State $\mathcal{F}' \gets \mathcal{F}' \cup \{f_k\}$
        \EndFor
    \EndFor
    \State $\mathcal{F} \gets \mathcal{F}'$;\quad $d \gets d + 1$
\Until{$d_{\texttt{max}} \leq d$} {\color{blue}{\Comment{\texttt{depth check}}}}
\end{algorithmic}
\end{algorithm}

\section{Simulations and Results}

\subsection{Experimental Setup}

We evaluated our framework on daily financial and macroeconomic data from 2020 to 2025 (Train: 2020–2023, Validation: 2024, Test: 2025).
We compared our model against a standalone LSTM.
We evaluated all models with the state-directional accuracy ($\textbf{Acc}_{\texttt{state}}$) and the return-sign accuracy (${\textbf{Acc}}_{\texttt{ret}}$) relative to the zero-return boundary.
$\textbf{Acc}_{\texttt{state}}$ measures how often the model correctly predicts the overall market state (\emph{bullish} or \emph{bearish}). 
$\textbf{Acc}_{\texttt{ret}}$ measures how often it correctly predicts the sign of the next day's return.
The dataset for Framework II simulation spans 119 days.
We evaluated the model on the data from February--July 2026.
The FCM for Framework II links each node to a ticker and masks the central node to avoid data leakage.

\subsection{Node Definitions}
Table~\ref{tab:nodes} lists the 16 concept nodes used in Framework I.
The table summarizes the 16 concept feature nodes across five categories, including Market Benchmarks, Tech Anchors, Macroeconomic Indicators, Technical Momentum, and Market Microstructure.
We used these concepts to construct the Framework I neuro-symbolic model. 
Framework II starts with the target node and grows until it reaches $d_{\max}$.

\begin{table}[!t]
\caption{\small Node definitions for Framework I FCMs:  FCM-A (market sentiment) and FCM-B (sector momentum) share the nodes.}
\label{tab:nodes}
\centering
\small
\begin{tabular}{|l|l|}
\hline
\multirow{1.0}{*}{\textbf{Notation}} & \multirow{1.0}{*}{\textbf{Node name}} \\
\hline
${C}_{\texttt{AI}}$   & \footnotesize{Technology Sector}         \\
${C}_{\texttt{VIX}}$  & \footnotesize{Market Volatility}         \\
${C}_{\texttt{Ene}}$  & \footnotesize{Energy / Inflation}         \\
${C}_{\texttt{Def}}$  & \footnotesize{Deficits}                   \\
${C}_{\texttt{Unemp}}$  & \footnotesize{Unemployment}               \\
\textcolor{black}{${C}_{\texttt{BI}}$}  & \footnotesize{\textcolor{black}{Business Investment}} \\
${C}_{\texttt{Sent}}$ & \footnotesize{Sentiment (dual-source)}    \\
\textcolor{black}{${C}_{\texttt{RSI}}$} & \footnotesize{\textcolor{black}{RSI-14 (overbought / oversold)}} \\
\textcolor{black}{${C}_{\texttt{M5}}$}  & \footnotesize{\textcolor{black}{Price Momentum (5-day)}}         \\
\textcolor{black}{${C}_{\texttt{M20}}$} & \footnotesize{\textcolor{black}{Price Momentum (20-day)}}       \\
${C}_{\texttt{qqq}}$  & \footnotesize{Market index} \\
${C}_{\texttt{nasdaq}}$  & \footnotesize{Target index} \\    
${C}_{\texttt{T3}}$  & \footnotesize{3-Month Treasury Bill Yield Rate} \\    
${C}_{\texttt{B5}}$  & \footnotesize{5-Year Breakeven Inflation Rate} \\ 
${C}_{\texttt{breadth}}$  & \footnotesize{Market Concentration Ratio} \\
${C}_{\texttt{tech-rot}}$  & \footnotesize{Sector Capital Rotation Differential} \\
\hline
\end{tabular}
\vspace{-0.1in}
\end{table}

\begin{figure*}[!t]
\centering
\begin{subfigure}[t]{0.44\textwidth}
\centering
\resizebox{\linewidth}{!}{%
\begin{tikzpicture}[scale=0.8,
  enode/.style={circle, draw=blue!65!black, fill=blue!9, thick,
                minimum size=1.0cm, font=\scriptsize\bfseries,
                align=center, inner sep=1pt},
  inode/.style={circle, draw=teal!75!black, fill=teal!9, thick,
                minimum size=1.0cm, font=\scriptsize\bfseries,
                align=center, inner sep=1pt},
  tnode/.style={circle, draw=orange!85!black, fill=orange!14, thick,
                minimum size=1.0cm, font=\scriptsize\bfseries,
                align=center, inner sep=1pt},
  snode/.style={circle, draw=violet!70!black, fill=violet!9, thick,
                minimum size=1.0cm, font=\scriptsize\bfseries,
                align=center, inner sep=1pt},
  nasdaq/.style={circle, draw=yellow!55!orange, fill=yellow!18,
                 line width=2.2pt, minimum size=1.55cm,
                 font=\small\bfseries, align=center},
  pos/.style={-{Stealth[length=5pt 1.6,width=3pt ]}, blue!55!black, line width=1.35pt},
  neg/.style={-{Stealth[length=5pt 1.6,width=4pt]}, red!65!black,   line width=1.35pt},
  posedge/.style={-{Stealth[length=5pt 1.6,width=3.2pt]}, blue!100!black, line width=1.35pt},
  negedge/.style={-{Stealth[length=5pt 1.6,width=4pt]}, red!100!black,  line width=1.35pt},
]
\node[nasdaq] (NQ) at (0,0) {\texttt{NASDAQ}};
\node[enode] (AI)  at ( 90.00:4.6) {AI\\Tech.\\ {\tiny\color{blue!50!black}$0.022$}};
\node[enode] (BI)  at ( 72.86:4.6) {Biz\\Invest.\\ {\tiny\color{blue!50!black}$0.015$}};
\node[enode] (CD)  at ( 55.71:4.6) {Consumer\\Demand \\ {\tiny\color{blue!50!black}$0.019$}};
\node[enode] (EPI) at ( 38.57:4.6) {Energy\\Price\,$\uparrow$ \\ {\tiny\color{red!70!black}$-0.053$}};
\node[enode] (SI)  at ( 21.43:4.6) {Supply\\Incr.\\ {\tiny\color{blue!50!black}$0.011$}};
\node[enode] (LU)  at (  4.29:4.6) {Low\\Unempl.\\ {\tiny\color{blue!50!black}$0.026$} };
\node[enode] (FD)  at (-12.86:4.6) {Foreign\\Demand \\ {\tiny\color{blue!50!black}$0.014$}};
\node[enode] (LD)  at (-30.00:4.6) {Large\\Deficits\\ {\tiny\color{red!70!black}$-0.021$}};
\node[enode] (MV)  at (-115.71:4.6){Market\\Volat.\\ {\tiny\color{blue!50!black}$0.030$}};
\node[enode] (EI)  at (-132.86:4.6){Equity\\Inflows\\{\tiny\color{blue!50!black}$0.038$}};
\node[inode] (IVS) at (-47.14:4.6) {Investor\\Sent.\\ {\tiny\color{blue!50!black}$0.028$} };
\node[inode] (CE)  at (-64.29:4.6) {Corp.\\Earn.\\ {\tiny\color{blue!50!black}$0.037$}};
\node[inode] (INF) at (-81.43:4.6) {Inflation \\ {\tiny\color{red!70!black}$-0.047$}};
\node[inode] (MP)  at (-98.57:4.6) {Monetary\\Policy \\ {\tiny\color{red!70!black}$-0.040$}};
\node[tnode] (RSI) at (-150.00:4.6){RSI-14 \\ {\tiny\color{red!70!black}$-0.035$}};
\node[tnode] (M5)  at (-167.14:4.6){Mom.\\5-day \\ {\tiny\color{blue!50!black}$0.033$}};
\node[tnode] (M20) at ( 175.71:4.6){Mom.\\20-day \\ {\tiny\color{blue!50!black}$0.019$} };
\node[snode] (NSY) at ( 158.57:4.6){Sent.\\Yahoo \\ {\tiny\color{blue!50!black}$0.017$} };
\node[snode] (NSG) at ( 141.43:4.6){Sent.\\Google \\ {\tiny\color{blue!50!black}$0.021$}};
\node[snode] (SA)  at ( 124.29:4.6){Sent.\\Agree. \\ {\tiny\color{blue!50!black}$0.024$} };
\node[snode] (SD)  at ( 107.14:4.6){Sent.\\Diverg. \\ {\tiny\color{red!70!black}$-0.016$} };
\draw[posedge] (AI)  -- (NQ);
\draw[posedge] (BI)  -- (NQ);
\draw[posedge] (CD)  -- (NQ);
\draw[posedge] (SI)  -- (NQ);
\draw[posedge] (LU)  -- (NQ);
\draw[posedge] (FD)  -- (NQ);
\draw[posedge] (IVS) -- (NQ);
\draw[posedge] (CE)  -- (NQ);
\draw[posedge] (EI)  -- (NQ);
\draw[posedge] (NSY) -- (NQ);
\draw[posedge] (NSG) -- (NQ);
\draw[posedge] (SA)  -- (NQ);
\draw[posedge] (MV)  -- (NQ);
\draw[posedge] (M5)  -- (NQ);
\draw[posedge] (M20) -- (NQ);
\draw[neg] (EPI) -- (NQ);
\draw[neg] (LD)  -- (NQ);
\draw[neg] (INF) -- (NQ);
\draw[neg] (MP)  -- (NQ);
\draw[neg] (RSI) -- (NQ);
\draw[neg] (SD)  -- (NQ);
\end{tikzpicture}%
}
\caption{Yahoo Finance FCM after online adaptation.}
\label{fig:yahoo_fcm}
\end{subfigure}
\hfill
\begin{subfigure}[t]{0.44\textwidth}
\centering
\resizebox{\linewidth}{!}{%
\begin{tikzpicture}[scale=0.8,
  enode/.style={circle, draw=blue!65!black, fill=blue!9, thick,
                minimum size=1.0cm, font=\scriptsize\bfseries,
                align=center, inner sep=1pt},
  inode/.style={circle, draw=teal!75!black, fill=teal!9, thick,
                minimum size=1.0cm, font=\scriptsize\bfseries,
                align=center, inner sep=1pt},
  tnode/.style={circle, draw=orange!85!black, fill=orange!14, thick,
                minimum size=1.0cm, font=\scriptsize\bfseries,
                align=center, inner sep=1pt},
  snode/.style={circle, draw=violet!70!black, fill=violet!9, thick,
                minimum size=1.0cm, font=\scriptsize\bfseries,
                align=center, inner sep=1pt},
  nasdaq/.style={circle, draw=yellow!55!orange, fill=yellow!18,
                 line width=2.2pt, minimum size=1.55cm,
                 font=\small\bfseries, align=center},
  pos/.style={-{Stealth[length=5pt 3.6,width=4pt]}, blue!55!black, line width=1.35pt},
  neg/.style={-{Stealth[length=5pt,width=4pt]}, red!65!black,   line width=1.35pt},
  posedge/.style={-{Stealth[length=5pt 1.6,width=3.2pt]}, blue!100!black, line width=1.35pt},
  negedge/.style={-{Stealth[length=5pt 1.6,width=4pt]}, red!100!black,  line width=1.35pt},
]
\node[nasdaq] (NQ) at (0,0) {\texttt{NASDAQ}};
\node[enode] (AI)  at ( 90.00:4.6) {AI\\Tech.\\{\tiny\color{blue!50!black}$0.020$}};
\node[enode] (BI)  at ( 72.86:4.6) {Biz\\Invest.\\{\tiny\color{blue!50!black}$0.013$}};
\node[enode] (CD)  at ( 55.71:4.6) {Consumer\\Demand\\{\tiny\color{blue!50!black}$0.018$}};
\node[enode] (EPI) at ( 38.57:4.6) {Energy\\Price\,$\uparrow$\\{\tiny\color{red!70!black}$-0.051$}};
\node[enode] (SI)  at ( 21.43:4.6) {Supply\\Incr.\\{\tiny\color{blue!50!black}$0.012$}};
\node[enode] (LU)  at (  4.29:4.6) {Low\\Unempl.\\{\tiny\color{blue!50!black}$0.024$}};
\node[enode] (FD)  at (-12.86:4.6) {Foreign\\Demand\\{\tiny\color{red!70!black}$-0.012$}};
\node[enode] (LD)  at (-30.00:4.6) {Large\\Deficits\\{\tiny\color{red!70!black}$-0.019$}};
\node[enode] (MV)  at (-115.71:4.6){Market\\Volat.\\{\tiny\color{blue!50!black}$0.028$}};
\node[enode] (EI)  at (-132.86:4.6){Equity\\Inflows\\{\tiny\color{blue!50!black}$0.036$}};
\node[inode] (IVS) at (-47.14:4.6) {Investor\\Sent.\\{\tiny\color{red!70!black}$-0.017$}};
\node[inode] (CE)  at (-64.29:4.6) {Corp.\\Earn.\\{\tiny\color{blue!50!black}$0.035$}};
\node[inode] (INF) at (-81.43:4.6) {Inflation\\{\tiny\color{red!70!black}$-0.045$}};
\node[inode] (MP)  at (-98.57:4.6) {Monetary\\Policy\\{\tiny\color{red!70!black}$-0.038$}};
\node[tnode] (RSI) at (-150.00:4.6){RSI-14\\{\tiny\color{red!70!black}$-0.033$}};
\node[tnode] (M5)  at (-167.14:4.6){Mom.\\5-day\\{\tiny\color{blue!50!black}$0.031$}};
\node[tnode] (M20) at ( 175.71:4.6){Mom.\\20-day\\{\tiny\color{red!70!black}$-0.022$}};
\node[snode] (NSY) at ( 158.57:4.6){Sent.\\Yahoo\\{\tiny\color{red!70!black}$-0.015$}};
\node[snode] (NSG) at ( 141.43:4.6){Sent.\\Google\\{\tiny\color{blue!50!black}$0.023$}};
\node[snode] (SA)  at ( 124.29:4.6){Sent.\\Agree.\\{\tiny\color{blue!50!black}$0.020$}};
\node[snode] (SD)  at ( 107.14:4.6){Sent.\\Diverg.\\{\tiny\color{red!70!black}$-0.014$}};
\draw[posedge] (AI)  -- (NQ);
\draw[posedge] (BI)  -- (NQ);
\draw[posedge] (CD)  -- (NQ);
\draw[posedge] (SI)  -- (NQ);
\draw[posedge] (LU)  -- (NQ);
\draw[posedge] (CE)  -- (NQ);
\draw[posedge] (EI)  -- (NQ);
\draw[posedge] (NSG) -- (NQ);
\draw[posedge] (SA)  -- (NQ);
\draw[posedge] (MV)  -- (NQ);
\draw[posedge] (M5)  -- (NQ);
\draw[neg] (EPI) -- (NQ);
\draw[neg] (FD)  -- (NQ);
\draw[neg] (LD)  -- (NQ);
\draw[neg] (IVS) -- (NQ);
\draw[neg] (INF) -- (NQ);
\draw[neg] (MP)  -- (NQ);
\draw[neg] (RSI) -- (NQ);
\draw[neg] (NSY) -- (NQ);
\draw[neg] (SD)  -- (NQ);
\draw[neg] (M20) -- (NQ);
\end{tikzpicture}%
}
\caption{Google-adapted FCM (4 edges differ from Yahoo).}
\label{fig:google_fcm}
\end{subfigure}
\caption{\footnotesize The numbers on each node indicate the edge weight link to the target.  (a) FCM edge polarities to the NASDAQ target after online adaptation on NASDAQ Composite data (2020--2025), shown in a spoke layout to eliminate clutter. Blue arrows denote excitatory influence ($w_{i,\texttt{NQ}}>0$); red arrows denote inhibitory influence ($w_{i,\texttt{NQ}}<0$). Node colour encodes category: blue = economic/market, teal = intermediate inner nodes, orange = technical indicators, violet = sentiment.
(b) Google Finance FCM after independent online adaptation on NASDAQ Composite data (2020--2025), using Google Finance price data and Google News RSS sentiment.
The graph structure is identical to Figure~\ref{fig:yahoo_fcm} but 4 edge signs differ (Foreign Demand, Investor Sentiment, Yahoo Sentiment, 20-day Momentum), reflecting divergent weight trajectories learned from a distinct data stream.
Both FCMs are adapted in isolation; their predictions are averaged with the LSTM in the ensemble forecast.
}
\label{fig:fcm_adapt}
\end{figure*}
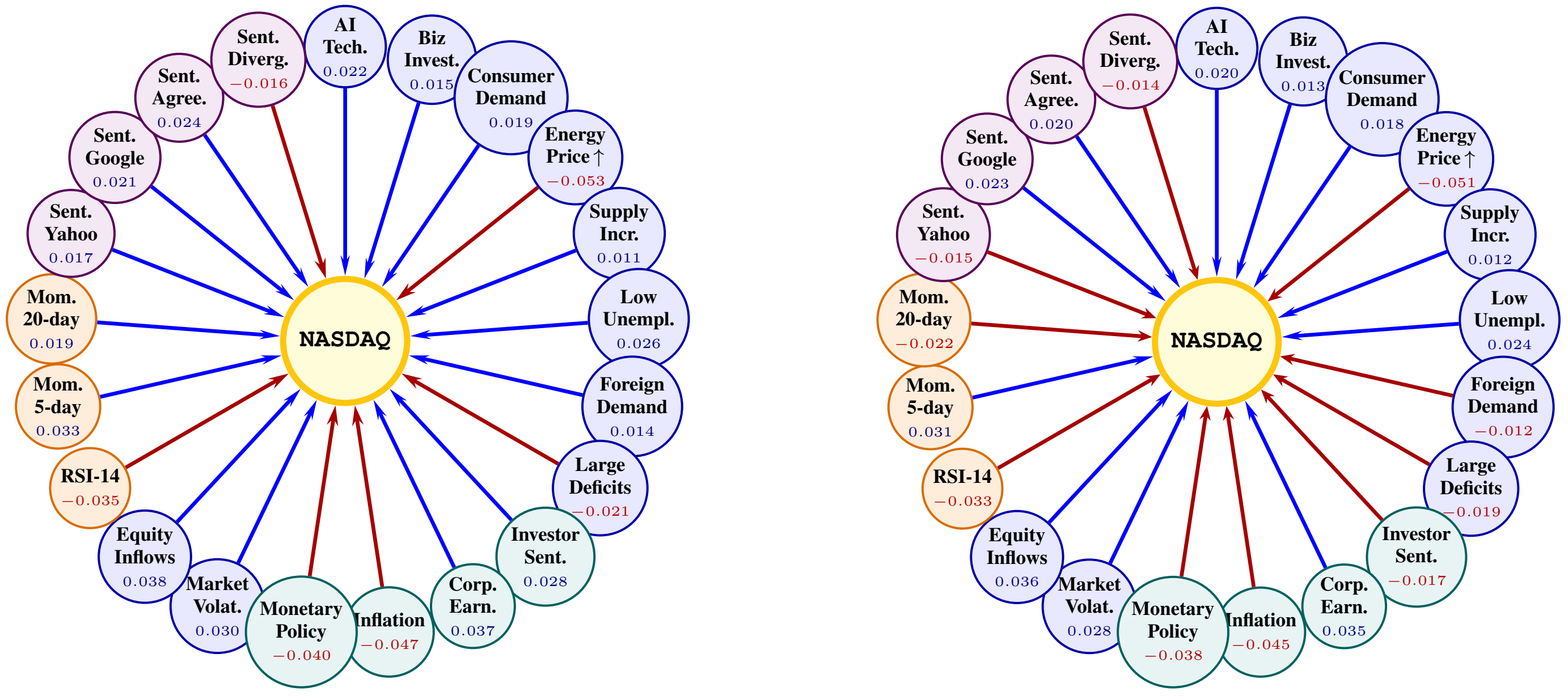

\begin{figure*}[!t]
\centering
\tikzset{
  anode/.style={circle, draw=blue!60!black, fill=blue!9, thick,
                minimum size=1.05cm, font=\scriptsize\bfseries, align=center},
  nqcirc/.style={circle, draw=yellow!55!orange, fill=yellow!18,
                 line width=2pt, minimum size=1.2cm,
                 font=\scriptsize\bfseries, align=center},
}
\begin{tikzpicture}[scale=0.68,
  posA/.style={-{Stealth[length=3.5pt 1.6]}, blue!100!black},
  negA/.style={-{Stealth[length=3.5pt 1.6]}, red!100!black},
]
\node[nqcirc] (NQ) at (0,0) {NQ};
\node[anode] (Ene) at ( 90:2.75) {Energy\\Price\,$\uparrow$\\ {\tiny\color{red!70!black}$-0.010$}}; 
\node[anode] (EI)  at ( 18:2.75) {Equity\\Inflows \\ {\tiny\color{blue!50!black}$0.008$}};
\node[anode] (MP)  at (-54:2.75) {Monetary\\Policy \\{\tiny\color{red!70!black}$-0.008$}}; 
\node[anode] (MV)  at (-126:2.75){Market\\Volat.\\ {\tiny\color{blue!50!black}$0.012$}};
\node[anode] (CE)  at ( 162:2.75){Corp.\\Earn. \\ {\tiny\color{blue!50!black}$0.010$}};
\draw[negA, line width=2.12pt] (Ene) -- (NQ)
      node[midway, above right, font=\tiny, black] {};
\draw[posA, line width=2.12pt] (EI)  -- (NQ)
      node[midway, right,       font=\tiny, black] {};
\draw[negA, line width=2.12pt] (MP)  -- (NQ)
      node[midway, below right, font=\tiny, black] {};
\draw[posA, line width=2.12pt] (MV)  -- (NQ)
      node[midway, below left,  font=\tiny, black] {};
\draw[posA, line width=2.12pt] (CE)  -- (NQ)
      node[midway, above left,  font=\tiny, black] {};
\node[font=\small\bfseries, anchor=north] at (0,-3.55)
      {$t = 0$ \textit{(Initial)}};
\end{tikzpicture}%
\hfill
\begin{tikzpicture}[scale=0.68,
  posB/.style={-{Stealth[length=3.5pt 1.6]}, blue!100!black},
  negB/.style={-{Stealth[length=3.5pt 1.6]}, red!100!black},
]
\node[nqcirc] (NQ) at (0,0) {NQ};
\node[anode] (Ene) at ( 90:2.75) {Energy\\Price\,$\uparrow$ \\ {\tiny\color{red!70!black}$-0.031$}};
\node[anode] (EI)  at ( 18:2.75) {Equity\\Inflows\\  {\tiny\color{blue!50!black}$0.020$}};
\node[anode] (MP)  at (-54:2.75) {Monetary\\Policy \\ {\tiny\color{red!70!black}$-0.025$}};
\node[anode] (MV)  at (-126:2.75){Market\\Volat.\\  {\tiny\color{blue!50!black}$0.019$}};
\node[anode] (CE)  at ( 162:2.75){Corp.\\Earn.\\  {\tiny\color{blue!50!black}$0.022$}};
\draw[negB, line width=2.12pt] (Ene) -- (NQ)
      node[midway, above right, font=\tiny, black] {};
\draw[posB, line width=2.12pt] (EI)  -- (NQ)
      node[midway, right,       font=\tiny, black] {};
\draw[negB, line width=2.12pt] (MP)  -- (NQ)
      node[midway, below right, font=\tiny, black] {};
\draw[posB, line width=2.12pt] (MV)  -- (NQ)
      node[midway, below left,  font=\tiny, black] {};
\draw[posB, line width=2.12pt] (CE)  -- (NQ)
      node[midway, above left,  font=\tiny, black] {};
\node[font=\small\bfseries, anchor=north] at (0,-3.55)
      {$t = 500 \,\mathrm{days}$ \textit{(Adapting)}};
\end{tikzpicture}%
\hfill
\begin{tikzpicture}[scale=0.68,
  posC/.style={-{Stealth[length=3.5pt 1.6]}, blue!100!black},
  negC/.style={-{Stealth[length=3.5pt 1.6]}, red!100!black},
]
\node[nqcirc] (NQ) at (0,0) {NQ};
\node[anode] (Ene) at ( 90:2.75) {Energy\\Price\,$\uparrow$ \\ {\tiny\color{red!70!black}$-0.053$}};  
\node[anode] (EI)  at ( 18:2.75) {Equity\\Inflows \\ {\tiny\color{blue!50!black}$0.038$}};
\node[anode] (MP)  at (-54:2.75) {Monetary\\Policy \\ {\tiny\color{red!70!black}$-0.040$}};
\node[anode] (MV)  at (-126:2.75){Market\\Volat.\\ {\tiny\color{blue!50!black}$0.030$}};
\node[anode] (CE)  at ( 162:2.75){Corp.\\Earn. \\ {\tiny\color{blue!50!black}$0.037$}};
\draw[negC, line width=2.12pt] (Ene) -- (NQ)
      node[midway, above right, font=\tiny, black] {};
\draw[posC, line width=2.12pt] (EI)  -- (NQ)
      node[midway, right,       font=\tiny, black] {};
\draw[negC, line width=2.12pt] (MP)  -- (NQ)
      node[midway, below right, font=\tiny, black] {};
\draw[posC, line width=2.12pt] (MV)  -- (NQ)
      node[midway, below left,  font=\tiny, black] {};
\draw[posC, line width=2.12pt] (CE)  -- (NQ)
      node[midway, above left,  font=\tiny, black] {};
\node[font=\small\bfseries, anchor=north] at (1.0,-3.55)
      {$t = 1250\,\mathrm{days}$ \textit{(Final)}};
\end{tikzpicture}
\caption{\footnotesize Online adaptation of FCM edge weights to the NASDAQ target node over the 2020--2025 training window. Five representative nodes are shown for clarity. The full adapted network appears in Figures~\ref{fig:yahoo_fcm}--\ref{fig:google_fcm}.
The numbers on each node indicate the edge weight link to the target.
\textbf{Left} ($t=0$, initialization): weights start from the 30-day sign-correlation prior (Eq.~\ref{eqn:weight_init}). All magnitudes are near zero.
\textbf{Center} ($ t= 500$ days, $\sim$2022): energy price and monetary policy establish clear inhibitory influence as the Fed tightening cycle begins. Equity inflows and corporate earnings grow excitatory.
\textbf{Right} ($t=1250$ days, final): energy price reaches the strongest inhibitory weight ($w=-0.053$). Market volatility stabilizes at $w=+0.030$, a regime-rebound signature.
Blue marks excitatory edges. Red marks inhibitory edges.\vspace{-0.1in}}
\label{fig:fcm_adaptation}
\end{figure*}

\subsection{Adaptation and Causal Dynamics}
{\color{black}{The online adaptation rule (Algorithm~\ref{alg:fcm_adapt}) updates the incoming edge weights to the target node at every trading day using the one-step prediction error.
Figure~\ref{fig:fcm_adaptation} shows the weight trajectories across the full 2020--2025 training window for five representative nodes.
The energy price increase node carries the strongest inhibitory effect on NASDAQ ($w = -0.053$), consistent with the negative relationship between commodity inflation and technology-sector valuations.
Deficits ($w = -0.021$) apply additional downward pressure, reflecting rising interest rates and equity risk premiums.
Conversely, market volatility stabilizes at $w = 0.03$, capturing the rebound dynamics that follow volatility spikes.
The system stabilizes to each new market regime within $10$--$15$ trading days, consistent with the convergence properties of online gradient descent with weight clipping to $[-1,1]$.}}

\subsection{Predictive Performance: Framework I}
{\color{black}{Table~\ref{tab:ensemble_result} reports the directional accuracies for each component model and the ensemble.
We evaluated on the 2025 test set ($n_{\text{test}} = 250$ trading days).
A prediction is classified as directionally correct when $\operatorname{sign}(\hat{r}_t - 0.5) = \operatorname{sign}(r_t - 0.5)$, where $\hat{r}_t$ is the predicted value.
The 0.5 threshold separates up from down days in the normalized return space.

\begin{table}[htbp]
\caption{\small Framework I Prediction: We measured the state-directional and return-sign accuracies (${\bf Acc}_{\texttt{state}}$ and ${\bf Acc}_{\texttt{ret}}$). 
The best steering parameters are $\alpha_{\text{up}} = 0.21$ and $\alpha_{\text{down}} =0.2$.}
\label{tab:ensemble_result}
\centering
\begin{tabular}{|l|c|c|}
\hline
\multirow{1}{*}{\bf Model} & \multirow{1}{*}{${\bf Acc.}_{\texttt{state}}$}  & \multirow{1}{*}{${\bf Acc.}_{\texttt{ret}}$}  \\
\hline
FCM A (Market-wide sentiment)           & $63.2\%$ & $48.8\%$ \\
FCM B (Tech. sector momentum)          & $63.2\%$ & $48.8\%$ \\
LSTM     & $78.4\%$ & $44.4\%$ \\
\textbf{Ensemble (FCMs and LSTM)} & {$ \bf 78.8\%$} & {$\bf 52.0\%$} \\
\hline
\end{tabular}
\end{table}

Our FCM-LSTM ensemble achieved $52.0\%$ return-sign accuracy, outperformed the standalone LSTM baseline ($44.4\%$) by $7.6$ percentage points, and outperformed the standalone FCMs ($48.8\%$) by $3.2$ percentage points. 
The ensemble matched the LSTM on state-directional accuracy ($78.8\%$ vs. $78.4\%$), while it outperformed pure FCM baselines ($63.2\%$) by $15.6$ percentage points.
We determined the asymmetric steering parameters ($\alpha_{\text{up}} = 0.210$ and $\alpha_{\text{down}} = 0.200$) with a two-dimensional grid search on the 2024 validation dataset to maximize directional accuracy while preserving neural calibration.
Standalone FCM-A (market-wide sentiment) and FCM-B (sector-specific momentum) produce closely aligned predictions ($r = 0.9998$) because \texttt{QQQ} represents over $80\%$ of \texttt{ONEQ's} market capitalization weighting, allowing dual-stream synthesis to filter single-stream noise while preserving tech sector sensitivity.

Table~\ref{tab:framework2_result} reports the state-directional accuracy for Framework II. 
It shows four variations of Framework II.
We evaluated the models by lagging input features by one trading day ($t-1$) to prevent look-ahead leakage. 
Term $\ell$ denotes the time-lag in days.
Same-day inputs ($\ell = 0$) reached $85\%$ accuracy.
The accuracy drops as the time-lag increases.

}

\begin{table}[t]
\centering
\caption{\small{Framework II Performance: State-directional accuracies across four knowledge sources with time-lag $\ell$ in days.
We evaluated the model on the data between February and July 2026 ($119$ days). }}
\label{tab:framework2_result}
\setlength{\tabcolsep}{4pt}
\renewcommand{\arraystretch}{1.25}
\begin{tabular}{|l|c|c|c|}
\hline
\textbf{FCM} & \textbf{$\ell=0$ } & \textbf{$\ell=1$}  &
\textbf{$\ell=7$}  \\
\hline
AI-Backed
  & $77.31\%$            & $50.42\%$ & $49.56\%$  \\
Research-Based
  & $\bf{85.71\%}$            & $50.42\%$ & $46.02\%$  \\
3-Layer
  & $68.91\%$            & $50.42\%$ & $\bf{50.44\%}$  \\
Stock-Based
  & $69.75\%$            & $\bf{54.62\%}$ & $\bf{50.44\%}$  \\
\hline
\end{tabular}
\end{table}

\vspace{-0.1in}
\section{Discussion}

Framework I shows a benefit of fusing FCM causal graphs with deep LSTMs for financial forecasting.
The framework uses the asymmetric additive residual fusion method to combine the two.
With the grid search over the validation set, the best steering values are $\alpha_{\text {up}} = 0.21$ and $\alpha_{\text {down}} = 0.20$.
This means that the ensemble permits a strong FCM conviction to adjust the LSTM prediction by up to $\pm 10.5\%$ in probability.
The slight asymmetry reflects market tail skewness, granting bullish macro momentum marginally higher steering weight during market rallies.
For Framework II, empirical results confirm that literature-grounded causal maps provide reliable same-day directional predictions.

\section{Conclusion}
We have presented a neuro-symbolic framework with two complementary parts.
Framework~I integrates dual adaptive FCMs, real-time news
sentiment, and an LSTM for financial market prediction.
This ensemble uses the asymmetric additive residual fusion to combine LSTMs and FCMs.
The ensemble outperformed the standalone LSTM and standalone FCMs.
Framework~II builds a literature-grounded, LLM-augmented map whose edges
carry verifiable citations, which makes its value the auditability of how
financial factors interact rather than raw forecasting. 
Together, the two
frameworks show that combining structural nonlinear dynamic modeling with deep learning yields better predictions and offers a promising direction for
adaptive and interpretable financial forecasting.

\bibliographystyle{IEEEtran}
\bibliography{references}

\vspace{12pt}
\end{document}